\documentclass[letterpaper]{article} 
\usepackage{aaai25}  
\usepackage{times}  
\usepackage{helvet}  
\usepackage{courier}  
\usepackage[hyphens]{url}  
\usepackage{graphicx} 
\usepackage{natbib}  
\usepackage{caption} 
\usepackage{amsfonts}  
\usepackage{amsmath}
\usepackage{subcaption}
\usepackage{booktabs}
\usepackage{multirow}
\usepackage{colortbl}
\usepackage[table]{xcolor}
\usepackage{tabularx}
\usepackage{soul}
\usepackage[switch]{lineno}
\usepackage{algorithm}
\usepackage{algorithmic}
\usepackage{dsfont}
\usepackage{newfloat}
\usepackage{listings}
\usepackage{fancyhdr}
\usepackage{xcolor}

\DeclareCaptionStyle{ruled}{labelfont=normalfont,labelsep=colon,strut=off} 
\floatstyle{ruled}
\newfloat{listing}{tb}{lst}{}
\floatname{listing}{Listing}
\nocopyright 

\title{SSNeRF: Sparse View Semi-supervised Neural Radiance Fields with Augmentation}

\author {
Xiao Cao\textsuperscript{\rm 1},
Beibei Lin\textsuperscript{\rm 1},
Bo Wang\textsuperscript{\rm 2},
Zhiyong Huang\textsuperscript{\rm 1},
Robby T. Tan\textsuperscript{\rm 1}
}
\affiliations {
\textsuperscript{\rm 1} National University of Singapore\\
\textsuperscript{\rm 2} CtrsVision\\
\{xiaocao, beibei.lin\}@u.nus.edu, \{dcshuang, robby.tan\}@nus.edu.sg, hawk.rsrch@gmail.com,
}

\begin{document}

\maketitle
\begin{abstract}\label{sec:abstract}
Sparse-view NeRF is challenging because limited input images lead to an under-constrained optimization problem for volume rendering. Existing methods address this issue by relying on supplementary information, such as depth maps. However, generating this supplementary information accurately remains problematic and often leads to NeRF producing images with undesired artifacts.
	To address these artifacts and enhance robustness, we propose SSNeRF, a sparse-view semi-supervised NeRF method based on a teacher-student framework. Our key idea is to challenge the NeRF module with progressively severe sparse-view degradation while providing high-confidence pseudo labels. This approach helps the NeRF model become aware of noise and incomplete information associated with sparse views, thus improving its robustness.
The novelty of SSNeRF lies in its sparse-view-specific augmentations and semi-supervised learning mechanism. In this approach, the teacher NeRF generates novel views along with confidence scores, while the student NeRF, perturbed by the augmented input, learns from the high-confidence pseudo-labels.
	Our sparse-view degradation augmentation progressively injects noise into volume rendering weights, perturbs feature maps in vulnerable layers, and simulates sparse-view blurriness. These augmentation strategies force the student NeRF to recognize degradation and produce clearer rendered views. By transferring the student’s parameters to the teacher, the teacher gains increased robustness in subsequent training iterations.
	Extensive experiments demonstrate the effectiveness of our SSNeRF in generating novel views with less sparse-view degradation. We will release code upon acceptance.
\end{abstract}

\section{Introduction}
{\let\thefootnote\relax\footnotetext{{Under Review}}}
\label{sec:intro}
Despite the rapid development of NeRF, they remain highly dependent on the quantity of input images \cite{yu2021pixelnerf,sensitive}. This dependency presents a significant challenge for deploying NeRF in real-world applications, where capturing a dense set of images is often impractical.

To address degradation in sparse-view settings, several methods have been proposed. Some approaches use supplementary information, such as depth maps~\cite{deng2022dsnerf,wang2023sparsenerf,roessle2022depth_prior}, object matching knowledge~\cite{truong2023sparf}, and pretrained generative models~\cite{jain2021dietnerf,diffusionnerf01}. These techniques tackle the challenge of limited data by incorporating external knowledge.
Other methods address the sparse-view NeRF problem through regularization rather than relying solely on external sources. These methods include constraining the distribution of sample points~\cite{seo2023mixnerf} and simulating bundle adjustment~\cite{lin2021barf}.

Methods that rely on external knowledge, such as depth maps, often yield suboptimal results, especially when the input scenes differ significantly from the training data~\cite{wang2023sparsenerf}. Regularization-based methods frequently struggle with complex scenes, such as those with rich textures or intricate geometry, due to their oversimplified assumptions, like a smooth loss or central-object assumption.
In contrast, we tackle sparse NeRF from three key perspectives: (1) enhancing NeRF's robustness to noisy ray densities, (2) making NeRF aware of and resilient to sparse-view blurriness, and (3) injecting noise into specific layers of our module to improve robustness against noisy input.

Building on these key perspectives, we introduce a teacher-student semi-supervised framework that incorporates sparse-view-specific augmentations.
The objective of our teacher NeRF is to generate pseudo labels for novel views with a high level of confidence. To prevent the confidence map from being biased towards high-contrast regions, we add some constraints in the HSV color space.
Once we obtain high-confidence regions, we generate paired data for novel views. By combining this with the original training data, we can effectively supervise NeRF, even in the presence of noisy, degraded augmentations.

\begin{figure*}[th]
	\centering
	\begin{tabular}{p{0.00\textwidth}p{0.20\textwidth}p{0.24\textwidth}p{0.24\textwidth}p{0.20\textwidth}p{0.00\textwidth}}
		& \centering\scriptsize MipNeRF &\centering\scriptsize  RegNeRF & \centering\scriptsize  MixNeRF& \centering\scriptsize  SSNeRF(Ours)&
	\end{tabular}
	\begin{subfigure}[b]{\textwidth}
		
		\includegraphics[width=\linewidth]{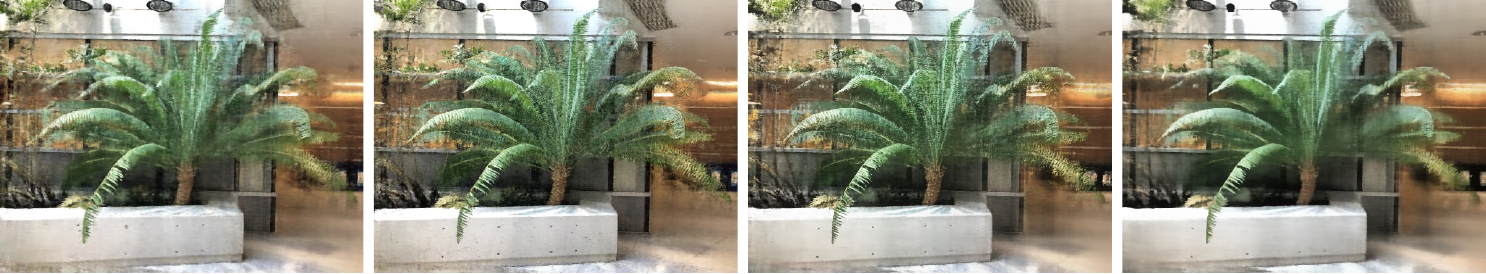}
		\caption{\scriptsize Fern 3-view}
		\label{fig:fern3v}
	\end{subfigure}

	\begin{subfigure}[b]{\textwidth}
		
		\includegraphics[width=\linewidth, height=3.7cm]{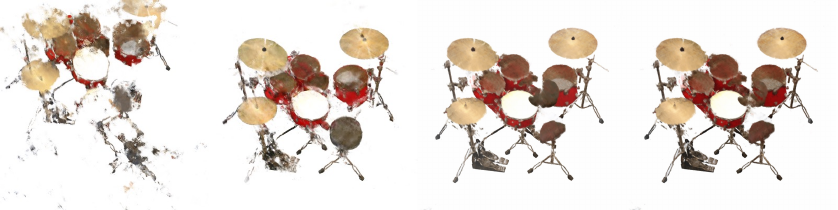}
		\caption{\scriptsize Drums 3-view}
		\label{fig:trex3v}
	\end{subfigure}

	\caption{
		Qualitative results on scene \textit{fern} and \textit{drums} under 3 training-view setting. Ours can effectively remove the hallucinations and floating points for rendered images. For zoom-in version, please refer to Figure~\ref{fig:zoom_in_visual}. Results of more scenes can be found in supplementary material.
	}
	\label{fig:compare_visual}
	
\end{figure*}

For the student learning process, we first inject noise into the weights of sampling points during volume rendering, enabling the student NeRF module to become aware of ray density noise.
After rendering new views, we apply a degradation simulation operation on this view to simulate the blurriness caused by sparse-view input.
Additionally, we observed that certain layers (i.e., vulnerable layers) of our NeRF module are particularly sensitive to the number of training views. To further enhance robustness of our student NeRF module, we perturb feature maps in those layers.

Having guidance from the teacher NeRF, which generates high-confidence pseudo-labels, we enable the student NeRF to recognize and adapt to sparse-view degradation.
To prevent NeRF from overfitting to a specific noise pattern, we transfer the student NeRF's generalized denoising capability back to the teacher NeRF. As a result, the teacher NeRF can render better-quality images and generate more accurate pseudo-labels for the student NeRF. Over time, the teacher NeRF develops a strong denoising ability.
As shown in Figure~\ref{fig:compare_visual} and Figure~\ref{fig:zoom_in_visual}, our method effectively handles the floating points and hallucinations in rendered images and videos (see in supplementary materials).

Our contributions can be summarized as follows:
\begin{itemize}
	
	\item  
	We design a novel semi-supervised NeRF framework (i.e., SSNeRF) that performs effectively in sparse-view settings. As a result, this framework enables NeRF to produce relatively sharp, artifact-free images.
	
	\item 
	In SSNeRF, sparse-view-specific augmentations are introduced, focusing on three aspects: 1) ray densities, 2) layer robustness, and 3) sparse-view blurriness.
    By incorporating high-confidence pseudo data pairs generated by teacher NeRF and augmentations in student NeRF, we enhance NeRF’s robustness to sparse-view noise.
	
	\item We conduct experiments on the real world complex dataset \textit{llff} and NeRF synthetic dataset \textit{blender} in different settings. 
	By gaining NeRF ability to recognize density-noise and sparse-view blurriness, SSNeRF could effectively remove the degradation of flickering pixels in rendered video.

\end{itemize}

\section{Related Work}

\paragraph{\textbf{Neural Radiance Field}}
NeRF~\cite{mildenhall2021nerf} is a neural network based on MLPs that predicts the RGB color and density of sampling points. It takes 5D input data (i.e., a 3D point $\textbf{x}=(x,y,z)$ and a view direction $\textbf{d}=(\Theta,\phi)$) and outputs the RGB color $c$ and density $\sigma$ of the sampled points in 3D space. These outputs are then rendered to the pixels using volumetric rendering. NeRF uses MSE as its loss function, meaning that there is no explicit supervision for densities in the 3D space, which contributes to the presence of noisy densities.
However, NeRF performs poorly in complex scenes~\cite{cao2024lightning}. MipNeRF~\cite{barron2021mipnerf}, a variant of NeRF, is known for its anti-aliasing capabilities achieved by casting cones instead of single rays. These enhancements allow MipNeRF to handle complex scenes with high-frequency details more effectively. Nevertheless, MipNeRF is designed for dense-view settings, and its performance decreases with fewer training views.

\paragraph{\textbf{Sparse View NeRF}}
Previous efforts to address the challenges of sparse-view NeRF can be categorized into two main approaches: 1) NeRF models incorporating depth information, and 2) NeRF models enhanced with pretraining information. A representative work in the first category is DSNeRF~\cite{deng2022dsnerf}, which uses sparse depth information generated by COLMAP~\cite{schoenberger2016colmap01,schoenberger2016colmap02} as an auxiliary supervision signal. Additionally, several studies have explored using dense depth information generated by neural networks. For example, Sparf~\cite{truong2023sparf} employs the point matching method PDC~\cite{truong2021PDC} to generate dense depth points, DDPNeRF~\cite{roessle2022depth_prior} uses depth completion to convert sparse depth into dense depth points, and SparseNeRF~\cite{wang2023sparsenerf} derives a relative depth map from DPT~\cite{ranftl2021dpt} to distill local depth rankings.
In the second category, DietNeRF~\cite{jain2021dietnerf} leverages CLIP ViT~\cite{radford2021clip} to extract semantic representations, transitioning from pixel-level to semantic-level supervision by incorporating 2D pretraining knowledge. NeRDi~\cite{deng2023nerdi} instead uses the Latent Diffusion Model (LDM)\cite{rombach2022LDM} to generate image priors. However, the performance deteriorates when the scene falls outside the scope of the prior knowledge. To address the instability associated with external knowledge, MixNeRF\cite{seo2023mixnerf} introduces a constraint on the density distribution along the cast rays. Despite these efforts to enhance stability, performance remains unsatisfactory, primarily due to the limited number of training views. These challenges highlight the need for strategies that more effectively utilize internal knowledge.

\paragraph{\textbf{NeRF Augmentation}}

Augmentation in NeRF can be implemented at three stages~\cite{augnerf}: 1) coordinate augmentation, 2) feature augmentation, and 3) pre-rendering output augmentation. Coordinate augmentation involves perturbing input camera matrix to compensate for inaccuracies in camera estimation. Feature augmentation targets MLP layers of NeRF, akin to traditional neural network augmentation techniques. Pre-rendering output augmentation adds noise to the RGB and density outputs. Although these methods are effective for dense-view NeRF, they are not well-suited for sparse-view settings and. Without sufficient supervision, some dense-view augmentations (e.g., camera pose perturbation) can lead to significantly blurry results.

\section{Proposed Method}\label{sec:method}
\begin{figure*}[ht!]
	\centering
	\includegraphics[width=\linewidth, height=0.47\textheight]{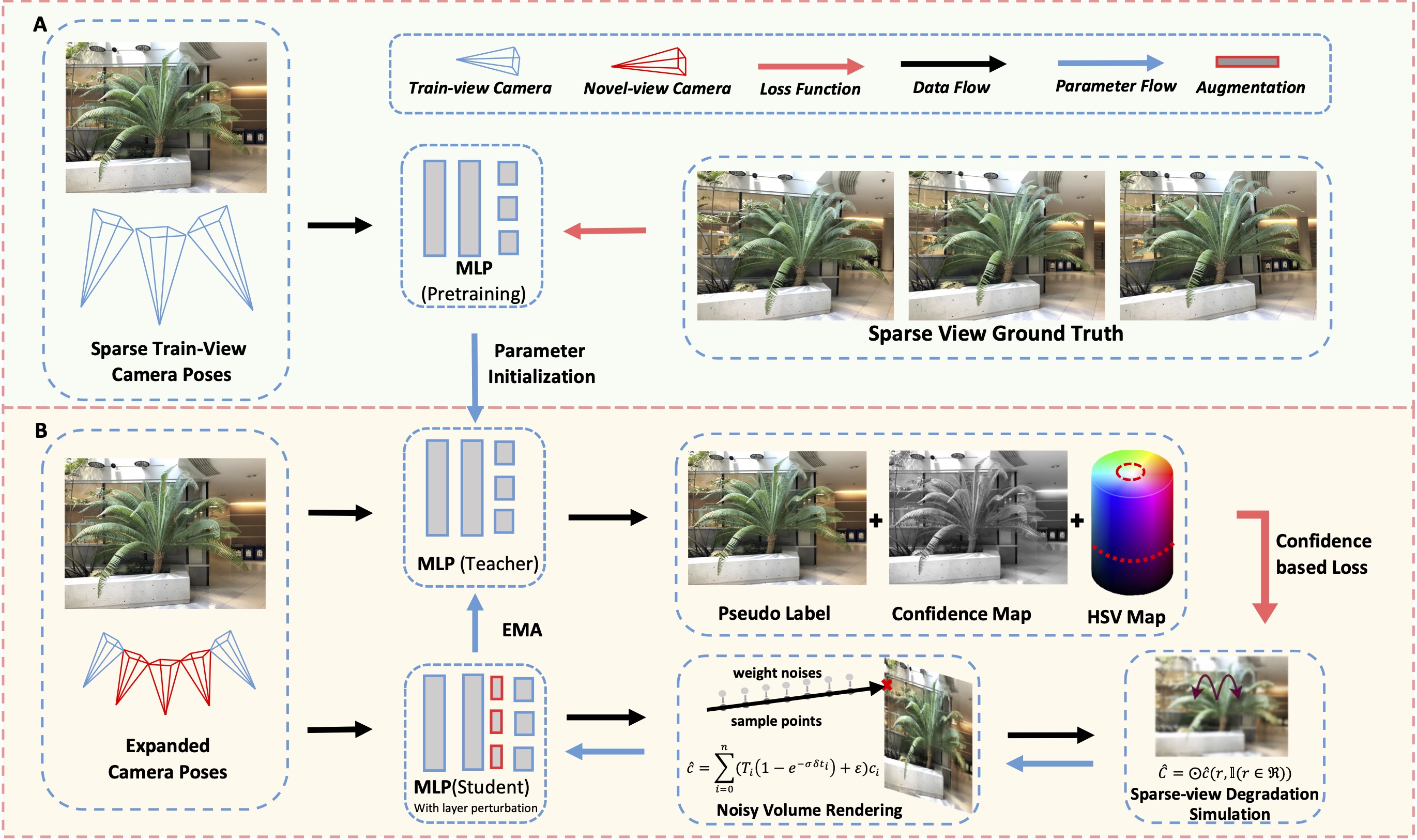}
	\caption{\textbf{SSNeRF Framework}. Our framework consists of two stages: (A) pretraining stage and (B) semi-supervised learning stage. The parameters are assigned to both teacher branch and student branch as initialization. Student branch is challenged with designed augmentation and supervised by teacher generated high-confidence pseudo-label together with sparse-view training data. The learned knowledge is passed to teacher by EMA, and at inference time, we remove all augmentations and only keep teacher NeRF.
	}
	\label{fig:framework}
\end{figure*}

Our method comprises two stages: a pretraining stage and a novel semi-supervised stage based on a teacher-student framework for NeRF, as shown in Figure~\ref{fig:framework}. 
In the first stage, we use labeled sparse-view training data pairs for pretraining. In the second stage, the framework generates unlabeled novel-view data pairs together with sparse-view training data to help NeRF overcome perturbations.

The second stage incorporates two key concepts: confidence map estimation with HSV constraints and sparse-view-specific augmentation. The novelty of this stage lies in simulating sparse-view degradation—such as noisy densities, sparse-view blurriness, and under-constrained layers—while guiding our student NeRF with high-confidence pseudo data and the original sparse-view training data. For inference, we use only the teacher NeRF from the SSNeRF as the final model (without the augmentation process)

\subsection{Semi-Supervised Teacher-Student}\label{sec:process}
The objective of our semi-supervised stage is twofold: first, to restore images generated from noisy sample-level densities to a quality comparable to those with high-quality sample-level densities, and second, to recognize and correct sparse-view blurriness. Through this process, our NeRF is able to predict adjusted sample-level RGB values in the presence of noisy densities and sparse-view blurriness. 
Those predicted adjusted sampling points’ RGB values, though not reflecting real 3D world colors if densities are entirely accurate, when combined with noisy densities, can still lead to accurate RGB values and solve the problem.

In our teacher-student framework, the teacher NeRF is responsible for generating confidence maps and pseudo ground-truths. Given the novel view camera pose matrix
$P_\text{novel}=[R|T]_{\rm{novel}}$,
the teacher NeRF predicts the pseudo ground truths $GT'$.
We then select the top $\kappa$ percent of high-confidence pixels as the pseudo ground truths.
We observe that high-confidence pixels selected by our epistemic confidence map are often concentrated in high-contrast regions (i.e., white and black pixels). This can lead NeRF to learn biased information, resulting in over-saturated images.
To mitigate this bias, we introduce an auxiliary HSV-based confidence map.
This map helps to select high-confidence pixels from the specific range of Saturation and Value of HSV,  and by combining two confidence maps together, we can obtain unbiased high-confidence pseudo labels.

\begin{figure*}[th!]
	\centering
	\begin{tabular}{p{0.00\textwidth}p{0.28\textwidth}p{0.12\textwidth}p{0.28\textwidth}p{0.12\textwidth}p{0.00\textwidth}}
		& \centering\scriptsize MixNeRF &\centering\scriptsize  Zoom-in Regions & \centering\scriptsize  Ours& \centering\scriptsize  Zoom-in Regions&
	\end{tabular}
	\begin{subfigure}[b]{0.95\textwidth}
		
		\includegraphics[width=\linewidth,height=3.4cm]{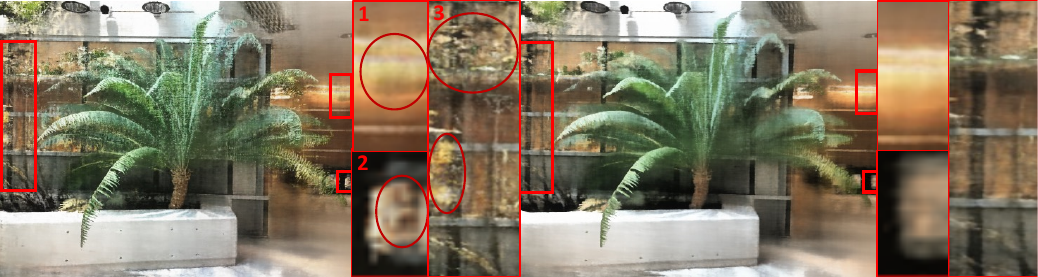}
		
		\caption{\scriptsize Fern 3-view (zoom-in)}
		\label{fig:zoom_fern}
	\end{subfigure}
	
	\begin{subfigure}[b]{0.95\textwidth}
		
		\includegraphics[width=\linewidth,height=3.4cm]{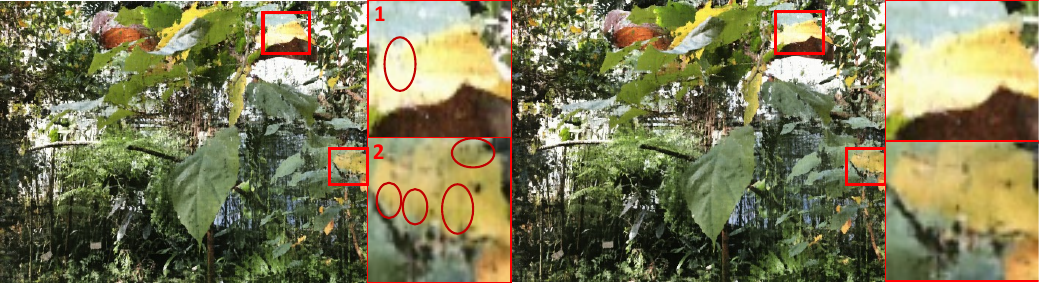}
		
		\caption{\scriptsize Leaves 3-view (zoom-in)}
		\label{fig:zoom_leaves}
	\end{subfigure}
	
	\begin{subfigure}[b]{0.95\textwidth}
		
		\includegraphics[width=\linewidth,height=3.4cm]{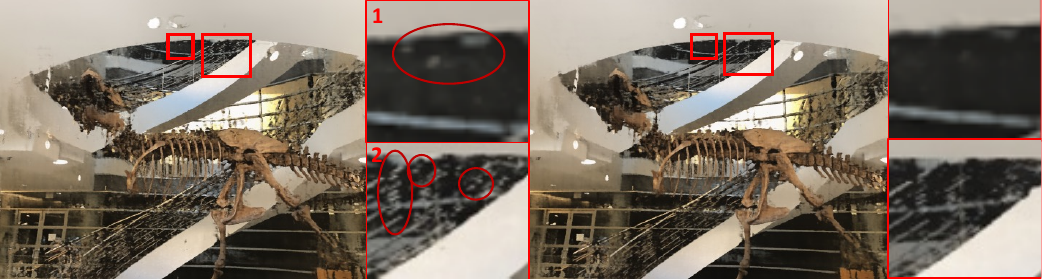}
		
		\caption{\scriptsize Trex 3-view (zoom-in)}
		\label{fig:zoom_trex}
	\end{subfigure}

	\caption{
		Zoom-in qualitative results on scene \textit{fern}, \textit{leaves}, and \textit{trex} with 3 training views. We select the regions that contain floating points with light red boxes and show the weakness by dark red circles. More can be found in supplementary materials.
	}
	\label{fig:zoom_in_visual}
	
\end{figure*}

\subsection{Unbiased Confidence Map}\label{sec:conf}
Our confidence map is an ensemble of the Monte-Carlo epistemic confidence map~\cite{dropout} $\tau_{e}$ and the HSV-confidence map $\tau_{v}$. This combination allows us to obtain the unbiased confidence map.
Epistemic uncertainty is used to estimate the model's uncertainty caused by lack of knowledge (i.e., models trained with insufficient data), which is suitable for the sparse view setting.
However, the confidence map generated by comparing outputs with and without dropout tends to be unstable and unreliable.

Instability arises because a single application of dropout can introduce significant randomness, leading to inconsistent uncertainty estimations. To address this, we use an ensembling strategy that applies a series of small, varied dropout ratios (e.g., 0\%, 5\%, 15\%, 20\%). By calculating the variance of each pixel across these different dropout configurations, the output variance reflects a broader range of potential outcomes rather than being overly influenced by a single, potentially extreme configuration~\cite{ensemble_dropout}. This approach results in a more stable and reliable confidence map. Based on our experimental results, the ensemble strategy achieves an average similarity ratio of approximately 93\% accuracy when compared with confidence maps generated from various combinations of dropout ratios.

We treat RGB channels independently and take the average as the final confidence score for each pixel. The averaged variance is viewed as a stable epistemic confidence map $\tau_{e}$. We select the top 10\% trustable pixels as our desired pseudo ground-truths.
However, as previously mentioned, the epistemic confidence estimator tends to select pixels from high-contrast regions because networks more easily learn extreme values compared to intermediate ones. This bias often leads to rendered images that are over-saturated and blurry.
To mitigate this, we introduce an additional confidence map estimation using the HSV color space.
We transform the aggregated dropout data into HSV, focusing on the Value and Saturation components.
By setting thresholds $v_{\rm{lower}}$ for \textbf{V} and $s_{\rm{lower}}$ for \textbf{S}, we filter out the dark and low-saturation regions.
These low-contrast regions are then used to re-estimate the epistemic confidence map. By combining both confidence maps, we can estimate the unbiased high-confidence regions in novel views.

\subsection{Sparse-View Specific Augmentation}
\label{sec:aug}

Synthesized novel view images in sparse-view settings often suffer from noisy geometry~\cite{niemeyer2022regnerf} (i.e., floating points) and blurry color. The presence of noise results in artifacts within the rendered images, flickering pixels and hallucinations in videos.
Although the predicted sample-level RGB values may be accurate, their combination with noisy sample point densities can still result in inaccurate pixel-level color. This leads to image generation characterized by blurriness (e.g., Figure~\ref{fig:zoom_fern}.1), floating points (e.g., Figure~\ref{fig:zoom_trex}.1), and hallucinations (e.g.,  Figure~\ref{fig:zoom_fern}.3).

Therefore, we need to equip our NeRF module with the ability to identify noise within the predicted sample-level densities and help NeRF also recognize the noise pattern caused by sparse-view input.
For this, we propose to finetune NeRF from three perspectives: a) noisy densities, b) vulnerable layers and c) sparse-view blurriness simulation.
From the noisy density perspective, we perturb the weights of sampling points to help NeRF recognize noise inherently from densities.

From vulnerable layers perspective, we augment the vulnerable layers caused by sparse-view settings to improve NeRF's robustness to noisy inputs. Note that the vulnerable layer here refers to the layer whose parameters change the most when changing the number of input views. 
From sparse-view blurriness simulation perspective, we further degrade the rendered result to simulate sparse-view blurriness and help NeRF become resilient to sparse-view blurriness.
Through these augmentations, NeRF is effectively challenged and gains ability to restore clear images despite noisy input.

We first conduct experiments with different sparse-view settings (i.e., different number of training views) to analyze the vulnerable layers (i.e., parameters of those layers change the most).
By calculating the variance of the parameters of the same layers across NeRF under different settings, we find that the ever final output layers (i.e., vanilla RGB and density output layers with additional output layers from backbone) in each module are the most sensitive.

Intuitively, perturbing input to layers can be viewed as deteriorating $N$-view NeRF to $(N-k)$-view NeRF and training to restore its reconstruction ability. And after this process, when removing the perturbation during inference time, the network becomes more robust. 
To this end, we design $\tau$-noise as in Equation~\ref{eq:noise} and inject into feature maps before vulnerable layers with progressively increased noise weight. Compared with dense-view settings feature augmenting, our method effectively enhances NeRF while reducing the training burden. 
As for density augmentation, we have two choices to severe the density noise and they will lead to different performance~\cite{ray_diffusion}: 1) perturb density directly and 2) perturb weight of sampling points (i.e., $weight=T_{i}\left(1-\exp \left(-\sigma_{i} \delta_{i}\right)\right)$) in the volume rendering process. Compared with perturbing weight in rendering process, perturbing density leads to more training costs due to one more mathematical transformation. Hence we choose to inject noise into the combination of transmittance and absorption coefficient, and then pass the denoising ability to teacher NeRF by EMA to prevent overfitting on a specific noise pattern~\cite{teacher_noise}. We design $\tau$-noise as following formulation for stable training purpose:

\begin{equation}
	\varepsilon \sim \mathbf{P}(x)=\frac{e^{e^{-x^2}}}{e^{e^{x^2}}+e^{e^{-x^2}}}.
	\label{eq:noise}
\end{equation}

For sparse-view degradation simulation, inspired by the dilation operation, we achieve the blurry effect by assigning the brightest color to nearby pixels, collaborating with a patch sampling strategy. The patch sampling strategy is applied only during the sparse-view simulation stage, as it reduces randomness (i.e., random patches with ordered pixels) and further exacerbates the overfitting problem. We apply the following operations:
%
%
\begin{table*}[ht!]
	\centering
	
	\setlength{\tabcolsep}{1.5mm}{
		\begin{tabular}{l|c|c|c|c|c|c|c|c|c}
			\toprule
			\multirow{2}{*}{Approach} & \multicolumn{1}{c}{PSNR $\uparrow$} & \multicolumn{1}{c}{SSIM $\uparrow$} & \multicolumn{1}{c}{LPIPS $\downarrow$} & \multicolumn{1}{c}{PSNR $\uparrow$} & \multicolumn{1}{c}{SSIM $\uparrow$} & \multicolumn{1}{c}{LPIPS $\downarrow$}& \multicolumn{1}{c}{PSNR $\uparrow$} & \multicolumn{1}{c}{SSIM $\uparrow$} & \multicolumn{1}{c}{LPIPS $\downarrow$}\\
			& 3-view & 3-view & 3-view & 4-view & 4-view & 4-view& 6-view    & 6-view    & 6-view   \\
			\midrule
			
			\multicolumn{10}{l}{\textit{\textbf{LLFF dataset}}} \\
			\hline
			
			MipNeRF  &\cellcolor{yellow!25} 17.377 &\cellcolor{yellow!25} 0.714  &\cellcolor{yellow!25} 0.335   &\cellcolor{orange!25} 19.142 &\cellcolor{yellow!25} 0.7852 &\cellcolor{yellow!25} 0.214   &\cellcolor{yellow!25} 22.122  &\cellcolor{yellow!25} 0.876   &\cellcolor{yellow!25} 0.150 \\
			RegNeRF  & 16.005 & 0.359  & 0.363   & 18.030 & 0.518  & 0.255    &  21.155 & 0.647   & 0.190\\
			MixNeRF  &\cellcolor{orange!25} 18.011 &\cellcolor{orange!25} 0.731  &\cellcolor{red!25} 0.243    &\cellcolor{yellow!25} 18.928 &\cellcolor{orange!25} 0.789  &\cellcolor{orange!25} 0.187    &\cellcolor{orange!25} 22.902 &\cellcolor{red!25} 0.891   &\cellcolor{orange!25} 0.132\\
			\hline
			\textbf{Ours} &\cellcolor{red!25}18.113&\cellcolor{red!25}0.748&\cellcolor{orange!25}0.250&\cellcolor{red!25}19.400&\cellcolor{red!25}0.796&\cellcolor{red!25}0.190&\cellcolor{red!25}22.990&\cellcolor{red!25}0.892&\cellcolor{red!25}0.110\\
			\hline\hline
			\multicolumn{10}{l}{\textit{\textbf{Blender dataset}}} \\
			\hline
			MipNeRF  & 10.316 & 0.445  & 0.385   & 9.812 & 0.397   & 0.494    &  18.844  & 0.875   & 0.166 \\
			RegNeRF  & \cellcolor{orange!25}18.791 & \cellcolor{yellow!25}0.763  & \cellcolor{orange!25}0.235   & \cellcolor{yellow!25}20.759 & \cellcolor{yellow!25}0.781  &\cellcolor{yellow!25} 0.277    &  \cellcolor{yellow!25}22.531 &\cellcolor{yellow!25} 0.817   & \cellcolor{yellow!25}0.227\\
			MixNeRF  & \cellcolor{yellow!25}18.420 & \cellcolor{orange!25}0.781  &\cellcolor{yellow!25} 0.250   & \cellcolor{red!25}21.821 & \cellcolor{orange!25}0.880  &\cellcolor{orange!25}0.209    &  \cellcolor{orange!25}25.289 & \cellcolor{orange!25}0.960   & \cellcolor{orange!25}0.151\\
			\hline
			\textbf{Ours} & \cellcolor{red!25}18.955 & \cellcolor{red!25}0.893 & \cellcolor{red!25}0.213 & \cellcolor{orange!25}21.357 & \cellcolor{red!25}0.925 & \cellcolor{red!25}0.153 & \cellcolor{red!25}25.391 & \cellcolor{red!25}0.974 & \cellcolor{red!25}0.096 \\
			
			\bottomrule
	\end{tabular}}
 \caption{\textbf{Quantitive comparison on LLFF and Blender dataset}. The reported results are performances under 3, 4 and 6 training views under dataset \textbf{\textit{llff}} and \textbf{\textit{blender}}. The \colorbox{red!25}{red}, \colorbox{orange!25}{orange} and \colorbox{yellow!25}{yellow} background refers to $1^{st}$, $2^{nd}$, $3^{rd}$ performance, respectively.}
 \label{tab:comparison}
\end{table*}
\begin{eqnarray}
	\mathbf{c}_i &=& f^{n} (r, \omega\mathds{1}(i=n)\varepsilon),\\
	\hat{c}(\mathbf{r}) &=& \sum_{i}^N (T_i (1 - e^{-(\sigma_i)\delta_i})+\omega\varepsilon)\mathbf{c}_i,\\
	\label{eq:render_aug}
	\hat{C}(\mathbf{r}) &=& \odot \hat{c}(\mathbf{r_{m},\mathds{1}(r\in \Re )}),
	\label{eq:render_aug}
\end{eqnarray}
where $f^{n}$ is the NeRF network with $n$ layers in each module and $\mathds{1}$ is the indicator function, $\varepsilon$ is the $\tau$-noise, $\omega$ is the progressively increased noise weight, $r_{m}$ is the ray with brightest color, $\Re$ is the rays within patch window and $\odot$ is the assigning function.

\subsection{Overall Objective Functions}
The overall loss function for the finetuning stage is:
\begin{equation}
	\rm{Loss}=\mathbb{L}(\hat{C}(\mathbf{r}),GT(x,y)_{\rm{pseudo}})_{_{(x,y) \in \tau}},
	\label{eq:fintune_loss01}
\end{equation}
where $\mathbb{L}$ is the original loss function in the pretraining stage.
We utilize MixNeRF as our pretrained backbone, and consequently, Equation~\ref{eq:fintune_loss01} can be formulated to Equation~\ref{eq:mixloss2}.:
\begin{equation}
	\mathcal{L}_\text{total} = \mathbb{L}_\text{MSE} + \lambda_{C}\mathbb{L}_\text{NLL}^{C} + \lambda_{D}\mathbb{L}_\text{NLL}^{D} + \hat{\lambda}_{C}\hat{\mathbb{L}}_\text{NLL}^{C},
	\label{eq:mixloss2}
\end{equation}
where $\mathcal{L}^{C}_\text{NLL}$ is the negative log-likelihood of the pdf of mixture model for sample points color, $\mathcal{L}^{D}_\text{NLL}$ is the negative log-likelihood of the pdf of mixture model for the depth of the $i$-th ray, $p(\mathbf{\tilde{c}}|\mathbf{r}_i)$ is negative log-likelihood of the pdf of mixture model for sample points color with regenerated depth~\cite{seo2023mixnerf}.

\section{Experiments}
In this section, we report quantitative results and qualitative results of baselines with teacher NeRF from SSNeRF. To save space, we put implementation details, quantitative results of ablation study, and analysis into supplementary material.

\paragraph{\textbf{Dataset}} To evaluate our proposed SSNeRF, we performed quantitative and qualitative evaluations on the LLFF dataset~\cite{mildenhall2019llff} and NeRF Synthetic Blender dataset~\cite{mildenhall2021nerf}.

\paragraph{\textbf{Metrics}} In our evaluation, we use Peak Signal-to-Noise Ratio (PSNR), Structural Similarity Index (SSIM) and Learned Perceptual Image Patch Similarity (LPIPS) as evaluation metrics, following sparse-view NeRF community~\cite{jain2021dietnerf,niemeyer2022regnerf,seo2023mixnerf,truong2023sparf,wang2023sparsenerf}.

\paragraph{\textbf{Baseline Methods}}
In our comparison experiment, we evaluate our model against several NeRF variants: RegNeRF~\cite{niemeyer2022regnerf}, a representative method of sparse-view NeRF; MixNeRF~\cite{seo2023mixnerf}, the state-of-the-art (SOTA) in knowledge-free sparse-view NeRF; and MipNeRF~\cite{barron2021mipnerf}, a leading model for dense-view NeRF. We utilize MixNeRF for pretraining process to obtain the basic reconstruction ability without introducing any external knowledge. All baselines are PyTorch reimplementation versions.

\subsection{Novel View Synthesis}
\label{sec:comparison}

\paragraph{\textbf{Quantitative Comparison}}
Table~\ref{tab:comparison} presents the quantitative comparisons of our method against three SOTA baseline methods. 
By aiding the external-knowledge-free method (i.e., MixNeRF), the performance improves steadily. For PSNR, the most improvement over the SOTA MixNeRF is 0.472 dB under \textit{llff} 4-view settings and 1.370dB compared with the famous RegNeRF.
Sparse-view RegNeRF performs worst in \textit{llff} dataset due to its ideal world assumption (i.e., central object and smooth surface). However, RegNeRF reaches better performances under \textit{Blender} dataset due to the dataset aligned with RegNeRF's assumption. This proves the unstable performance of constraining NeRF based on specific assumptions and further demonstrates the necessity of proposing a semi-supervised framework that frees NeRF from external prior.
By employing our semi-supervised mechanism, we can achieve the best performance in most cases and guarantee a top-three performance among all datasets.

\begin{table}[ht!]
	\centering

	\begin{tabular}{l|c|c|c}
		\toprule 
		Augmentation & PSNR $\uparrow$ & SSIM $\uparrow$ & LPIPS $\downarrow$  \\
		\midrule 
		Vanilla MixNeRF & 19.076 & 0.738 & 0.345  \\
		\hline
		Layer augmentation & 19.194 & 0.744 & 0.365  \\
		\hline
		Sparse-view Degradation & 19.282 & 0.0.745 & 0.365  \\
		\hline
		Weight Perturbation & 19.233 & 0.746 & 0.364  \\
		\hline
		Density Perturbation&19.208 &0.745 & 0.365  \\
  		\hline
		Ours(student)& 17.289 & 0.676 & 0.394  \\
		\hline
		Ours& 19.275 & 0.746 & 0.364  \\
		
		\bottomrule 
	\end{tabular}
 \caption{\textbf{Quantitative results of ablation studies}. The reported results are separated performances of each module under \textit{fern} 3-view setting}
\label{tab:ablation} 
\end{table}

\paragraph{\textbf{Qualitative Comparison}}
We visualize the qualitative results of all the baselines under 3 training view settings in Figure~\ref{fig:compare_visual} and zoom-in Figure~\ref{fig:zoom_in_visual} (more can be found in supplementary materials).
The qualitative results demonstrate the effectiveness of our method: images without hallucination and floating points (e.g., black dots on the surface of the plant and white points near the ceiling, as shown in~\ref{fig:zoom_in_visual}).
To better visualize the improvement, we visualize the zoom-in differences of SOTA MixNeRF with ours in Figure~\ref{fig:zoom_in_visual} to better prove our method's restoration ability. 

From the qualitative result, we observe that our SSNeRF renders the images with the least noise and preserves the most structural information. As shown in Figure~\ref{fig:compare_visual}, among all baselines, MipNeRF and RegNeRF render relatively noisy images. MipNeRF renders the most noisy images due to its dense-view setting. RegNeRF renders images with the most blurry high-frequency regions, performing even worse than the dense-view NeRF (i.e., MipNeRF). 
By comparing our results with SOTA MixNeRF in Figure~\ref{fig:zoom_in_visual}, ours achieve the clearest qualitative results with the least undesired hallucination.
\paragraph{\textbf{Discussion on Qualitative results}}
We highlight hallucinations with red boxes in Figure~\ref{fig:compare_visual} and notice two kinds of floating points: 1) floating points occurring at the wrong depth (manifest as hallucinations) and 2) floating points caused by the surroundings. For the first kind, take Figure~\ref{fig:zoom_fern} as an example, there are yellow hallucinations on the left side due to the inability to distinguish the background from the plant in the foreground. For the second kind, as seen in Figure~\ref{fig:zoom_trex}, the SOTA method MixNeRF renders white floating points around the white ceiling. When rendering images in the training view, it is hard to observe floating points because the noisy sampling points are along the same ray. However, in the novel view, these points lie on different rays, causing floating points in empty space. A similar situation is observed with the dustbin in Figure~\ref{fig:zoom_fern}, where black floating points appear in white regions, influenced by the black surroundings. Furthermore, since the sampling points with noisy densities lie on different rays in different views, the floating points appear in various positions in the rendered images. Combining such novel-view images into a video can result in flickering pixels. By eliminating these, our method can render the clearest video, free from flickering pixels (refer to videos in supplementary material).

\subsection{Ablation Studies}
\label{sec:ablation}
In this section, we report the quantitative result of using each augmentation separately within our semi-supervised framework in Table~\ref{tab:ablation}. The performance of each separate augmentation outperforms vanilla MixNeRF, demonstrating their effectiveness. Moreover, combining these augmentations results in an accumulated performance increase. Additionally, 'Weight perturbation' outperforms 'Density perturbation', and this further supporting our analysis in Section \textit{Sparse-view Specific Augmentation} that augmenting weight is more efficient than directly injecting noise into density due to one mathematical operation. To investigate the necessity of the teacher branch and student branch, we conducted experiments with only the student NeRF in the semi-supervised process. It performs the worst due to its overfitted denoising ability which we proved in the following experiments.

\begin{table}[h!]
	\centering

	\begin{tabular}{l|c|c|c}
		\toprule 
		Augmentation & PSNR $\uparrow$ & SSIM $\uparrow$ & LPIPS $\downarrow$  \\
		\midrule 
		MipNeRF & 18.571 & 0.703 & 0.493  \\
		\hline
		RegNeRF & 17.417 & 0.395 & 0.775  \\
		\hline
		MixNeRF & 18.727 & 0.712 & 0.412  \\	
  		\hline
		Ours(student)&17.569 &0.671 & 0.409  \\
		\hline
		Ours&18.793 &0.713 & 0.408  \\
		
		\bottomrule 
	\end{tabular}
 \caption{\textbf{Quantitative results of anti-noise ability}. The reported results are under \textit{fern} 3-view setting with uniform noise injected into density}
 \label{tab:uniform} 
\end{table}

\subsection{Analysis on Generalized Denoising Ability}
To further demonstrate the generalized denoising capability of SSNeRF, we inject uniform noise into the sampling-point densities. The results are presented in Table~\ref{tab:uniform}. We observe that baseline methods deteriorate severely under uniform noise, whereas our approach performs the best. In contrast, the student NeRF in the semi-supervised process performs the worst when exposed to uniform noise. This indicates that the student NeRF overfits to specific $\tau$-noise patterns and loses its generalized denoising ability. By employing EMA, our teacher NeRF effectively mitigates this overfitting problem and restores clear image effectively.

\section{Limitation}\label{sec:limitation}
We observe that SSNeRF improves less under \textit{"flower"} compared with MixNeRF. This is because the rendered image already achieves good quality and has few floating points. The failure case of SSNeRF happens when backbones are already capable of rendering clear images. And with the help of our finetuning framework, the performance improvement is trivial. For a few scenes, applying part of augmentations performs better than applying all. In addition, without the help of generation model, SSNeRF can not generate novel views of unseen regions.

\section{Conclusion}\label{sec:conclusion}

We aim to address the problem of NeRF sparse-view degradation in complex scenes. To tackle this difficulty, we propose SSNeRF, a novel teacher-student based novel-view learning framework for NeRF.
Our main novel idea lies in the semi-supervised process, which enables our NeRF model to render clear novel views even in the presence of noisy density maps.
We propose to use effective sparse-view specific augmentation from density, vulnerable layers and degradation simulation perspectives to empower NeRF with the capability to recover clear images without additional input. To prevent NeRF from learning biased knowledge, the ensembled epistemic confidence map is used together with HSV confidence map. With the aid of our novel semi-supervised process, we can render noise-free images and hallucination-free videos under sparse view settings.

\bigskip

\bibliography{aaai25.bib}
\end{document}